\documentclass[conference]{IEEEtran}       % onecolumn (second format)
\usepackage{graphicx}
\usepackage{amssymb}
\usepackage{flushend}

\makeatletter
\newcommand*{\rom}[1]{\expandafter\@slowromancap\romannumeral #1@}
\makeatother

\begin{document}

\title{Assessing fish abundance from underwater video using deep neural networks}

\author{
\IEEEauthorblockN{Ranju Mandal\IEEEauthorrefmark{1},  Rod M. Connolly\IEEEauthorrefmark{2}, Thomas A. Schlacher\IEEEauthorrefmark{3} and Bela Stantic\IEEEauthorrefmark{1}}

\IEEEauthorblockA{
\IEEEauthorrefmark{1}School of ICT, Griffith Sciences, Griffith University, QLD 4222, Australia 
}
\IEEEauthorblockA{
\IEEEauthorrefmark{2}Australian Rivers Institute - Coast \& Estuaries and \\ School of Environment and Science, Griffith University, QLD 4222, Australia\\ 
\{r.mandal, r.connolly, b.stantic\}@griffith.edu.au
}
\IEEEauthorblockA{
\IEEEauthorrefmark{3}School of Science and Engineering, University of the Sunshine Coast, QLD 4558, Australia
\\tschlach@usc.edu.au}
}
\maketitle

%\vspace{-3ex}

\begin{abstract}
Uses of underwater videos to assess diversity and abundance of fish are being rapidly adopted by marine biologists. Manual processing of videos for quantification by human analysts is time and labour intensive. Automatic processing of videos can be employed to achieve the objectives in a cost and time-efficient way. The aim is to build an accurate and reliable fish detection and recognition system, which is important for an autonomous robotic platform. However, there are many challenges involved in this task (e.g. complex background, deformation, low resolution and light propagation). Recent advancement in the deep neural network has led to the development of object detection and recognition in real time scenarios. An end-to-end deep learning-based architecture is introduced which outperformed the state of the art methods and first of its kind on fish assessment task. A Region Proposal Network (RPN) introduced by an object detector termed as Faster R-CNN was combined with three classification networks for detection and recognition of fish species obtained from Remote Underwater Video Stations (RUVS). An accuracy of 82.4\% (mAP) obtained from the experiments are much higher than previously proposed methods.


\end{abstract}
 \vspace{-1ex}
\section{Introduction}
\label{sec_intro}
%objectives
It is now common for marine scientists to assess fish abundance using multiple underwater video cameras \cite{Salman2016}. This innovative method of assessing fish populations is a viable alternative because it is inexpensive and non-lethal compared to traditional methods (i.e. uses of seine nets, fyke nets, gill nets, electrofishing, rotenone, and trawls) \cite{Wilson2015}. A Remote Underwater Video Stations (RUVS)-based approach can also work in complex habitats such as reefs or dense aquatic vegetation where traditional approaches are ineffective. Videos generated from RUVS are now mostly analyzed manually by fish taxonomy experts. These experts estimate fish abundance in different habitats to determine spatial patterns in fish abundance and species composition for a variety of research objectives. Information such as types of species and frequency of occurrence of a particular species are most important in this type of analysis.

\begin{figure*}[!htb]
   \centering
   \begin{tabular}{cc}
        \includegraphics[width=0.4\textwidth]{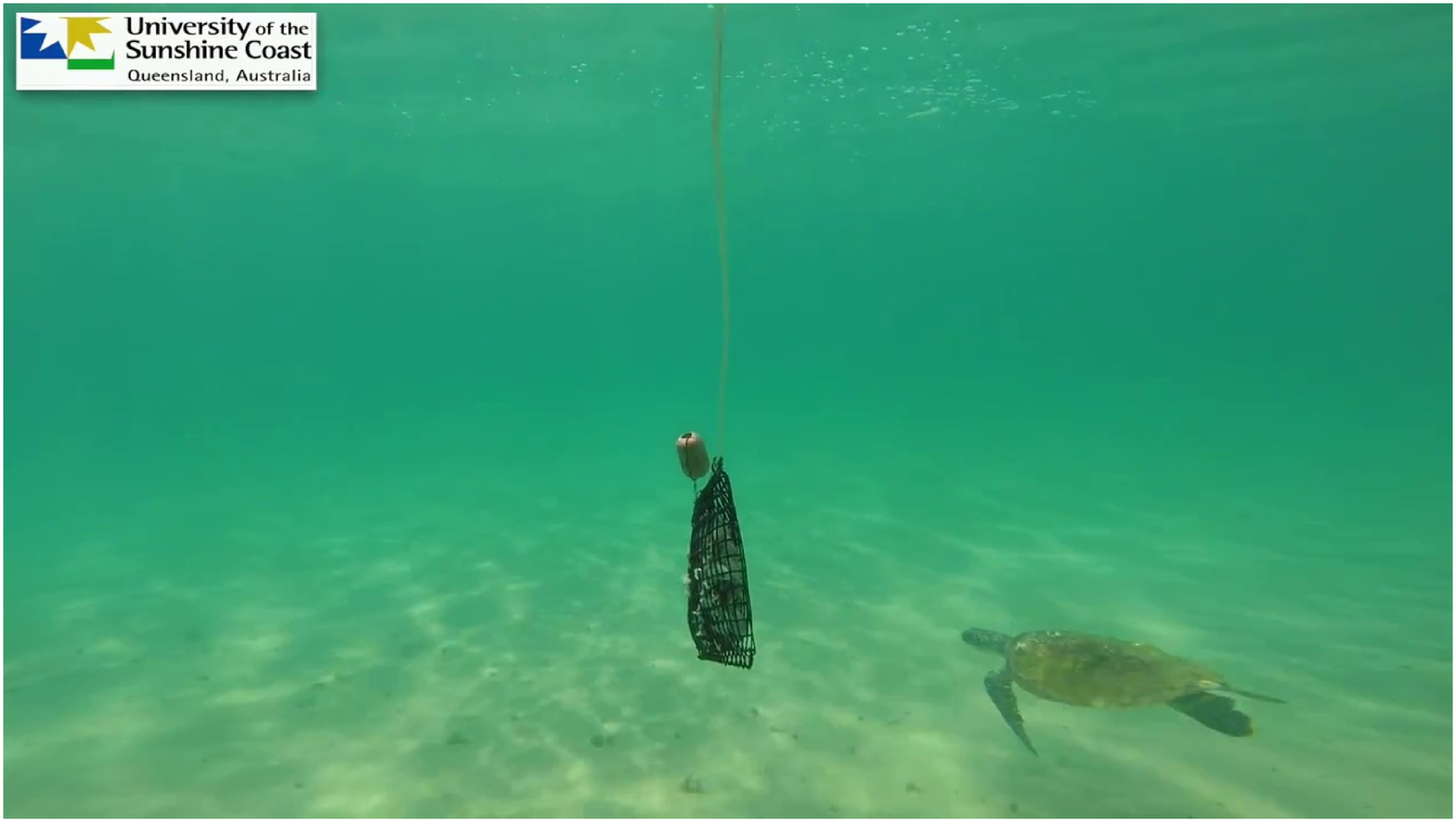}&
        \includegraphics[width=0.4\textwidth]{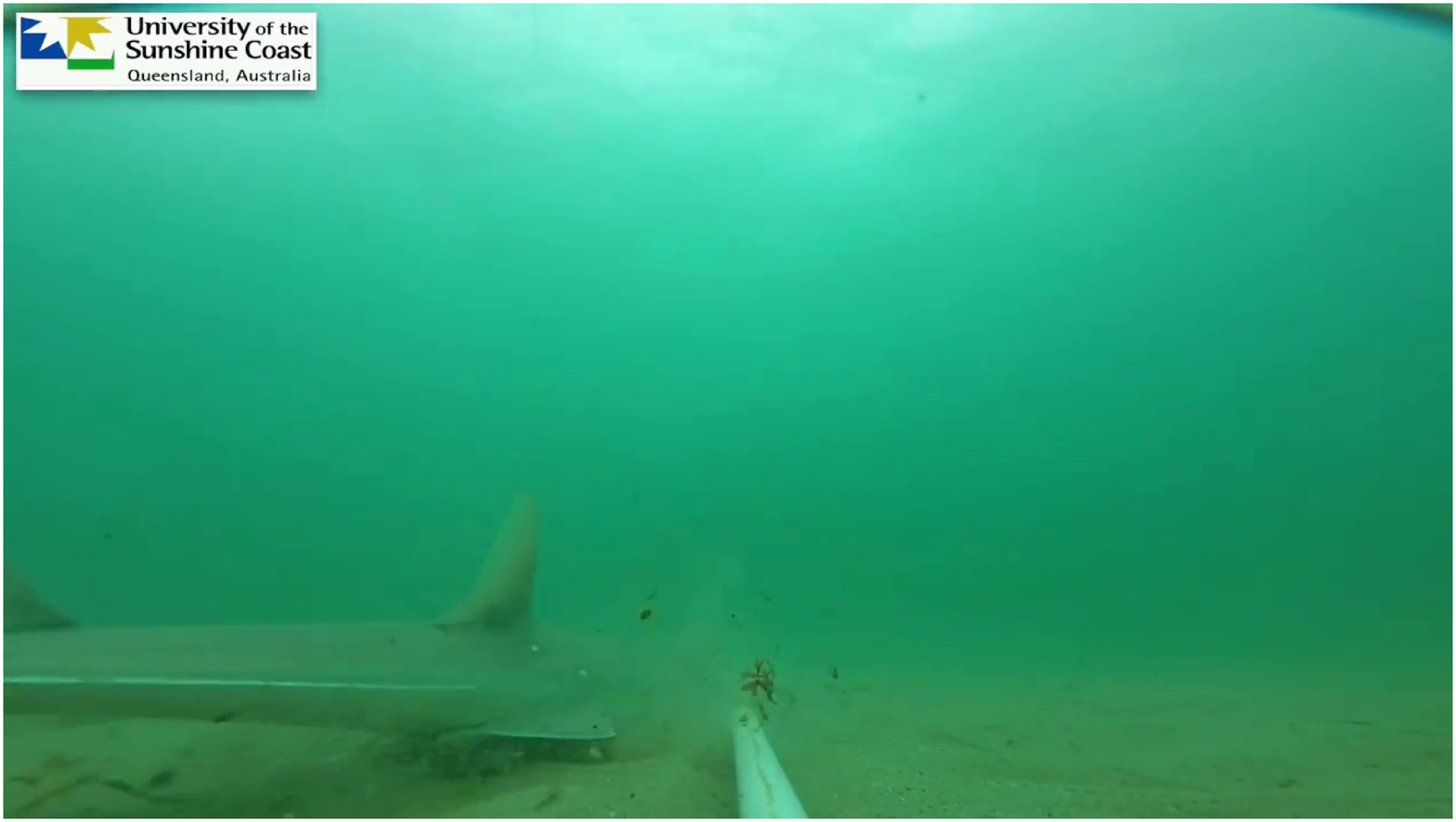}\\
        (a) & (b)\\
        \includegraphics[width=0.4\textwidth]{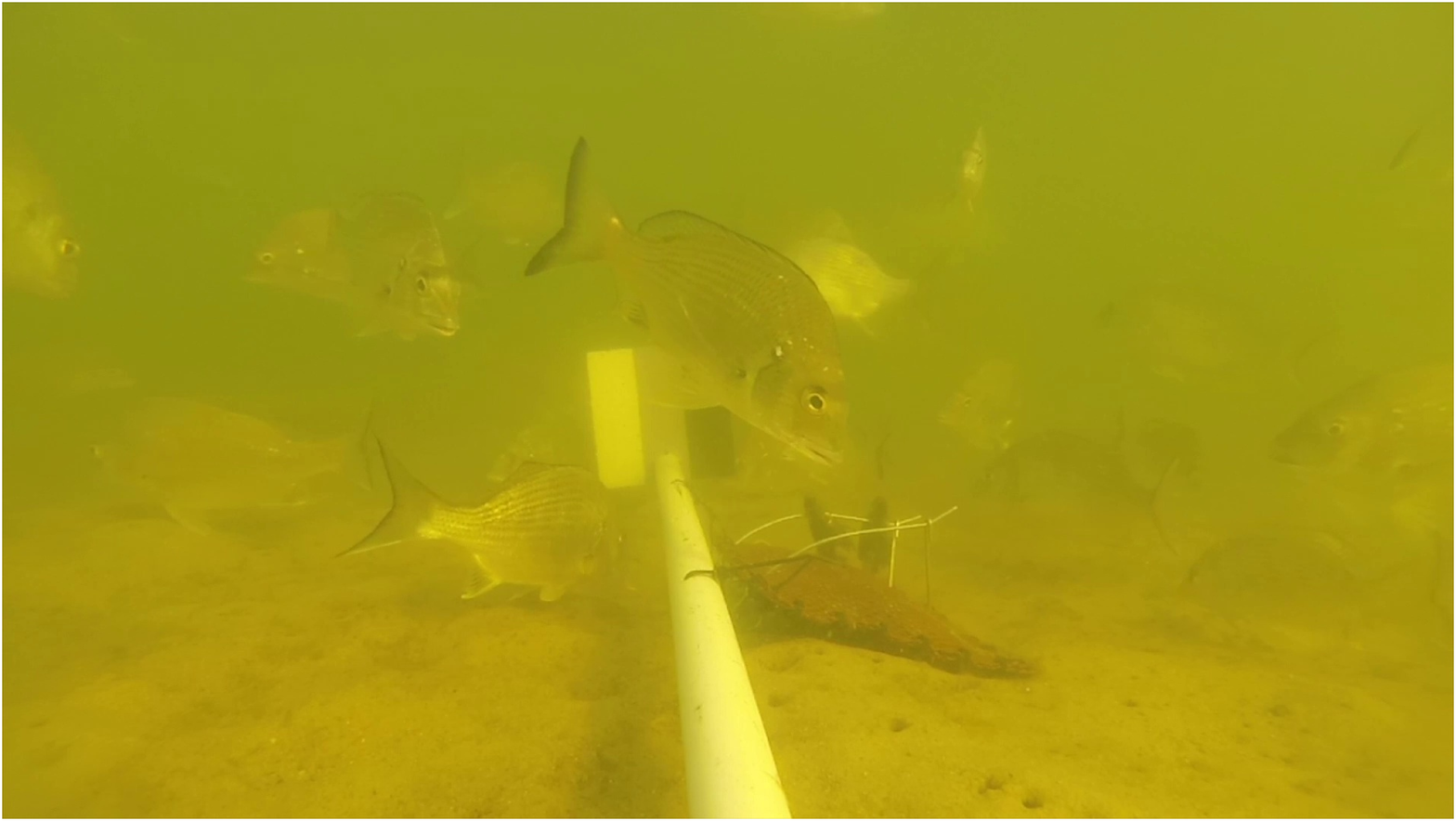}&
        \includegraphics[width=0.4\textwidth]{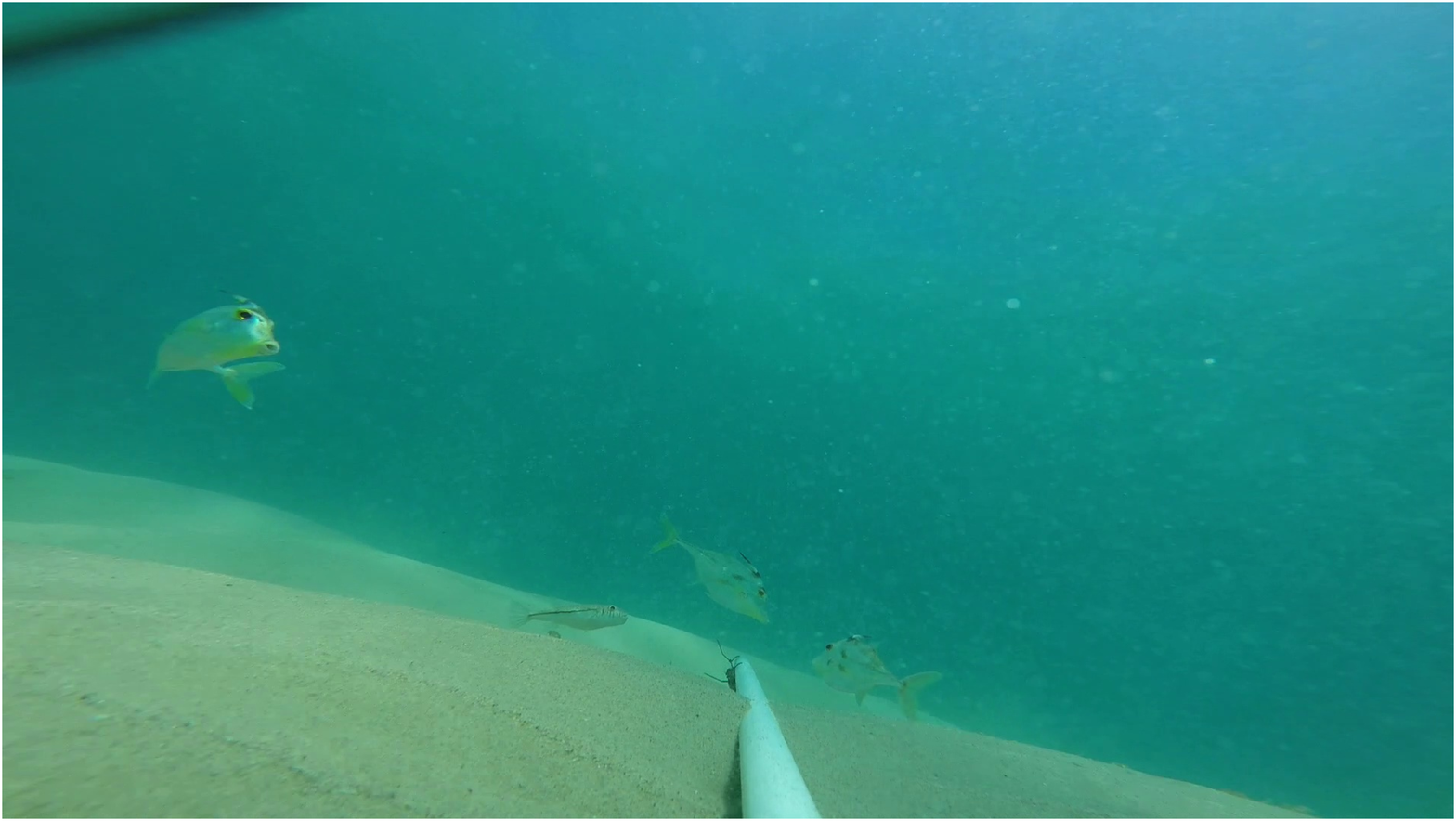}\\
        (c) & (d)\\
   \end{tabular}
   \caption{Sample frames extracted from underwater videos obtained from different beaches across southeast Queensland, Australia. (a,b) Coolum Beach  and (c,d) Currimundi Beach}
   \label{fig:SampleSurfFrames}
   \vspace{-4ex}
\end{figure*}

%%challenges
However, manual analysis of large amounts of video produced by clusters of RUVS is a tedious process and as it needs experts with specialized domain knowledge \cite{Gilby2k17}, the process becomes expensive. Automatic processing of captured underwater visual data from RUVS would be an ideal solution in such circumstances. Automatic detection of fish and other marine species is an essential step in order to distinguish the fish from the background (e.g. ocean floor, plants, rocks). This detection task is made more complex by the high levels of occlusion (due to schooling by fish), color and texture of fishes.  Fig.\ref{fig:SampleSurfFrames} shows some sample frames obtained from different marine sites across southeast Queensland, Australia.

Existing works in the literature are mostly semi-automatic \cite{Wilson2015, Spampinato2016} and assumed a constrained environment. An unconstrained video stream involves more complex environments and challenges like illumination, water turbidity, complex background, a variable number of species, changes in orientation and scale due to freely moving fishes. These factors pose a real challenge in recognition of species in an unconstrained environment. In this work, we looked to fully automate the process to obtain all the required information needed for an assessment from a captured underwater video. Two main components involved in this automation are (a) automatic detection of species bounding boxes in the frame and (b) classification of all the detected bounding boxes (Region of Interest) into predefined classes (i.e. species names). The proposed work addresses both challenges in a single pipeline using a deep learning-based end-to-end architecture called `Faster R-CNN'. 

The objective is to identify all the species present in an underwater video in a real-time scenario. The proposed architecture for fish assessment has many advantages apart from its fully automated properties. The system can successfully detect and recognise multi-oriented and multi-scale samples of species available in the dataset obtained from an unconstrained environment. A wide range of experiments was conducted using three different deep learning models and a dataset was developed with a significant number of surf species. 
%The paper is organized as follows. The related works on fish identification and recent deep learning approaches on object detection are discussed in section \ref{rworks}. The proposed method is presented in section \ref{method}. In section \ref{results}, experimental analysis and results are discussed. Finally, the paper is concluded in section \ref{conclusion}.
%\vspace{-1ex}
\section{Related works}
\label{rworks}
Despite significant literature for automatic object detection and recognition using deep learning, limited attention has been given to recognition of species from the underwater video for assessing fish abundance. We provide a brief review of the relevant research and state-of-the-art approaches on fish identification from underwater video footages. The limitations of the approaches in the literature are investigated to identify the gap and scope of works.  Existing methods can be categorised into two classes: handcrafted feature-based \cite{Rova07,Gundam2015,Spampinato2016} and machine learning-based \cite{Salman2016} approaches.
A deformable Template Matching-based feature extraction technique was proposed by Rova et al. \cite{Rova07} for classification of fish. Support Vector Machines (SVMs) with linear and polynomial kernel were used to classify fishes in their work.  Gundam et al. \cite{Gundam2015} proposed a fish classification technique which could be used as a partial automation of underwater video processing.  A Kalman filter-based technique was used. However, a constant velocity model was assumed which is not very compatible with the unpredictability of fish movements (velocity and directions). The shape-based feature (Fourier) extraction technique was employed which might not perform well when the number of classes increases and with fishes of identical shape.  Only three fish species were considered in the experiments whereas many more fish classes can be presents in undersea environments.\\
\indent Spampinato et al. \cite{Spampinato2016} proposed two different methods for fish detection in underwater images and videos contain ten different classes of fish. Three different approaches were proposed for image-based fish species recognition based on spooling and sparse coding-based features.  A two-step approach was adopted for fish detection and classification in videos.  A background subtraction-based approach was used to detect fishes whereas SIFT-descriptors and SVM-based classifier were used for recognition. However, limitations of shape context-based features and template matching techniques assume a constraint environment which is not applicable to real-time unconstrained underwater environment. Recently, the latest generation of Convolutional Neural Networks (CNNs) outperformed the approaches based on handcrafted approaches in computer vision research \cite{Chatfield14}. The problem of fish classification was addressed by Salman et al. \cite{Salman2016} using a CNN-based feature and SVM-based classifier.  The LifeCLEF fish dataset used in this experiment mainly contains fish templates.\\
% % goal and contribution
\indent Accurate object detection and classification still remain a challenging problem in the field of computer vision despite a significant progress being made using deep convolutional neural networks on image classification and detection \cite{ILSVRC15}. Recent advancement of deep ConvNets \cite{PierreICLR14OverFeat} has significantly improved the object detection and classification task. The object detection is a more challenging job, compared to image classification, as it requires more advanced and complex methods \cite{ GirshickCVPR2014-RCNN, PierreICLR14OverFeat} to obtain accuracy.  However, convolutional neural networks (CNNs) have now been successfully employed recently \cite{Uijlings13,edge-boxes}. The selective search \cite{Uijlings13} method merges superpixels based on low-level features and EdgeBoxes \cite{edge-boxes} uses edge information to generate region proposals, and these are now widely used. Shortcomings of proposed methods are that they need as much running time as the detection network to hypothesize object locations.\\
\indent Here, the recent state-of-the-art methods towards object detection \cite{GirshickCVPR2014-RCNN, girshickICCV15fastrcnn, renNIPS15fasterrcnn} has been discussed.  The Region-based Convolution Network Network (R-CNN) \cite{GirshickCVPR2014-RCNN} performs excellent object detection by using a deep ConvNet and classify the object proposals. R-CNN uses Selective Search (SS) technique to compute multi-scale object proposal to achieve the scale-invariance capability.  However, R-CNN is computationally expensive due to the processing of high numbers of object proposal and provides only rough localization which compromises speed and accuracy.\\ 
\indent Fast R-CNN \cite{girshickICCV15fastrcnn} is an improved version of R-CNN with a much faster training and testing process and it achieves more accuracy compare to R-CNN. R-CNN does not share computation and performs CovNet forward pass for each object proposal. Spatial pyramid pooling nets \cite{HeSPM-eccv} proposed a sharing computation technique which speeds up R-CNN but fine-tuning algorithm proposed in SPPnets \cite{HeSPM-eccv} cannot update the layers precede the Spatial pyramid pooling. In addition, as it deals with a variable window size of pooling, one stage (end to end) training was difficult. Fast-R-CNN fixes the drawbacks of R-CNN and SPPnet, whiling improving their speed and accuracy. The single-stage training process in Fast R-CNN can update all network layer using a multi-task loss and does not need disk storage for feature caching. 
In all of the above approaches, the power of CNN has been used only for regression and classification. The concept of Fast R-CNN was extended further in Faster R-CNN \cite{renNIPS15fasterrcnn} by introducing a Region Proposal Network (RPN).  The Faster R-CNN merges the RPN and Fast R-CNN into a single network by sharing their convolutional features using a popular terminology of neural networks with `attention' mechanisms, the RPN guides the network for object regions. RPN consists of several additional convolutional layers, build on top of the convolutional feature map.  Although the accuracy of R-CNN and Fast R-CNN were satisfactory, they were computationally expensive which make them unsuitable for real-time applications, unlike Faster R-CNN. We, therefore, selected the  Faster R-CNN \cite{renNIPS15fasterrcnn} as our approach in this investigation.
%Methodology starts here:
\section{Methodology}
\label{method}
\begin{figure*}[!htb]
    \centering
      \includegraphics[width=0.60\textwidth]{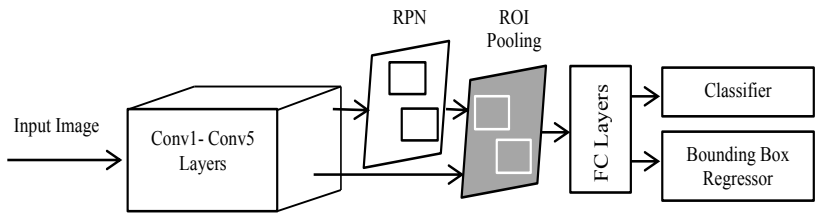}
    \caption{ An architecture of Faster R-CNN. Here, the feature map is generated by 5 layers convolution which is shared by the RPN network and the region classifier.}
   \label{fig:FasterRCNN}
   \vspace{-4ex}
\end{figure*}

The object detector called Faster R-CNN \cite{renNIPS15fasterrcnn} is a particularly successful method for general object detection. It is a single integrated network which consists of two modules: (a) region proposal, and (b) region classifier.  Fig. \ref{fig:FasterRCNN} shows a Faster R-CNN architecture which is a single, unified network for object detection. A deep fully convolutional network proposes a set of regions and then the regions are used by the Faster R-CNN \cite{girshickICCV15fastrcnn} detector. The Region Proposal Networks (RPNs) are designed to predict region proposals with a wide range of scales and aspect ratios. Sharing of convolution at test time with the very efficient object detection network \cite{girshickICCV15fastrcnn} significantly reduces the marginal cost of proposals computation.

The proposed RPN model \cite{renNIPS15fasterrcnn} can be combined with a classification model to achieve the detection and classification in an end to end framework. Different CNN-based classification models with different sizes (small, medium and large) were combined with the RPN network in our experiments to obtain Faster R-CNN models of three different sizes (i.e. the number of layers). The RPN consists of a few additional convolutional layers that simultaneously regress region bounds and objectness scores at each location on a regular grid. In faster R-CNN, RPN was constructed (see Fig. \ref{fig:FasterRCNN}) on top of the convolutional feature map which was trained end-to-end to generate high-quality region proposals. The following three classification models (ZF, CNN-M, and VGG-16) were used in our experiments to combine with RPN and compare the performances. 

{\bf ZF Net \cite{DBLP:journals/corr/ZeilerF13}:} Architecture of this network model is similar to AlexNet with minor modifications.  The filter size was reduced to $7 \times 7$  compared to $11 \times 11$ in AlexNet in the first convolutional layer of ZF net which helps to retain a significant pixel information in the input data. ZF net used ReLUs as activation functions, for error function cross-entropy loss and the network trained as batch stochastic gradient descent.

{\bf CNN-M \cite{Chatfield14}:} Architecture of this model is similar to the ZF model with some modifications. A smaller receptive field of the first convolutional layer and a decreased stride was shown to be beneficial. However, convolutional layer 2 uses a larger stride (2 instead of 1) to keep the computation time reasonable. The main difference between this model and  ZF is that CNN-M uses fewer filters in the layer 4 (512 instead of 1024).

{\bf VGG-16 Net \cite{Simonyan14c}:} This CNN model consists of 19 layers that only used $3 \times 3$ filters with a stride and pad of 1,  max-pooling with $2 \times 2$  and stride 2. The filter size of $3 \times 3$ is a contrast to ZF Nets $7 \times 7$ filter. An effective receptive field of $7 \times 7$ was achieved using 3 back to back conv layers. The model has used scale jittering as data augmentation during training and ReLU layers are used after each convolutional layer. The batch gradient descent was used during training.

The Caffe deep learning library \cite{jia2014caffe} was used for all the experiments presented here. In our experiments, publicly available pertained Caffe models for object detectors were used for initial weights and to enable transfer learning technique. Hence, to take advantage of all network architectures used in our experiments, transfer learning technique from ImageNet \cite{ILSVRC15} was used during fine-tuning of our models. A better performance and a faster convergence can be achieved using the transfer learning technique.

{\bf Implementation details:} All experiments have been conducted on an Intel(R) Xeon(R) CPU E5-2609 v3 @ 1.90GHz Linux cluster node with a GeForce GTX 1080 GPU installed. The python interface code was used to conduct all the experiments. The models are trained with a learning rate of 0.001 and batch size of 128. The RPN batch size was set to 256 for region-based proposal networks (RPN). Regions proposal networks were trained end-to-end using back-propagation and stochastic gradient descent (SGD). Non-maximum suppression (NMS) was employed to the proposals based on the class scores to reduce redundancies arising from RPN proposals. Performance of each network architecture at different iterations was also analyzed. In the training phase, the snapshot of trained models was saved at an interval of 10k iterations. Detections with overlap greater than the 50\% Intersection over Union (IoU) threshold with the corresponding ground-truth bounding box are considered as true positive and all other detections as false positive  and IoU calculated as:
%\vspace{-3ex}
\begin{equation}
\label{eq_iou}
IoU(BB_{pred}, BB_{gt}) = \frac{area(BB_{pred}\cap BB_{gt})}{area(BB_{pred} \cup BB_{gt})}
\end{equation}
where $BB_{pred}$ and $BB_{gt}$ denotes predicted bounding box and ground truth bounding box respectively. The Average Precision (AP) is computed for each class, while mean Average Precision (mAP) denotes the mean over all the computed APs.
%------------------------------------------------------------------------------
\section{Results and Discussions}
\label{results}
%---------------------------------------------------------------------
{\bf Dataset:} Details about the fish datasets used for the experiments are described here. Underwater videos used in our experiments were provided by the authors as part of a collaborative research program based at University of Sunshine Coast \cite{Borland2k17, GILBY2017132}. The videos contain fish communities in marine waters of beaches and estuaries across southeast Queensland, obtained using baited and unbaited GoPro cameras. 4909 images containing 12365 annotated samples of 50 species of fish and crustaceans were used in our experiments. The Vatic  interactive video annotation tool \cite{springerlink:10.1007/s11263-012-0564-1} was employed to annotate the data and was standardized in PASCAL VOC \cite{Everingham15} format. The dataset was divided into training, validation, and test sets using a random sampling technique. The training, validation and test set comprises of 70\%, 10\%, and 20\% data respectively.

\begin{figure*}[!htb]
   \centering
   \begin{tabular}{cc}
        \includegraphics[width=0.4\textwidth]{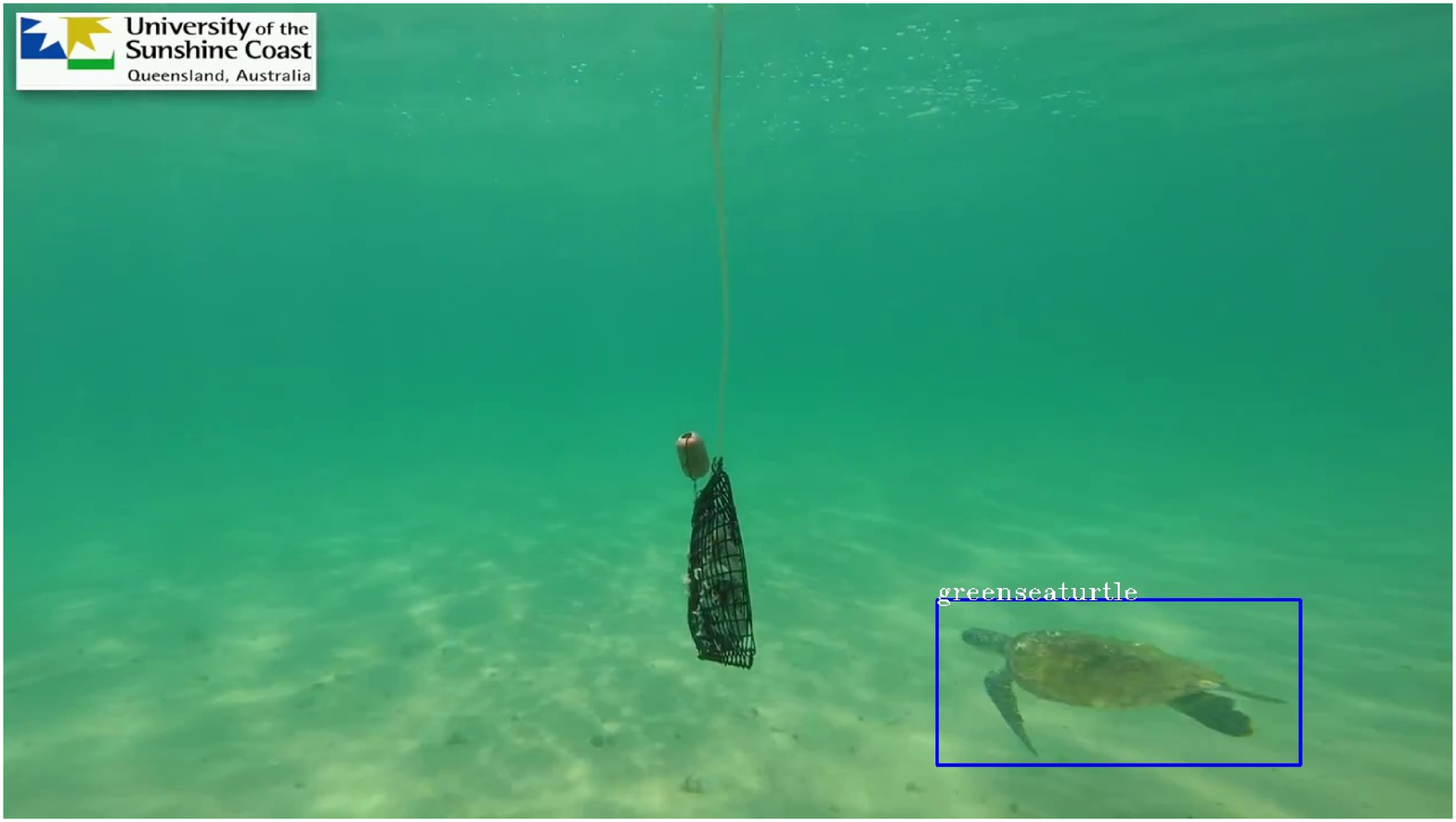}&
        \includegraphics[width=0.4\textwidth]{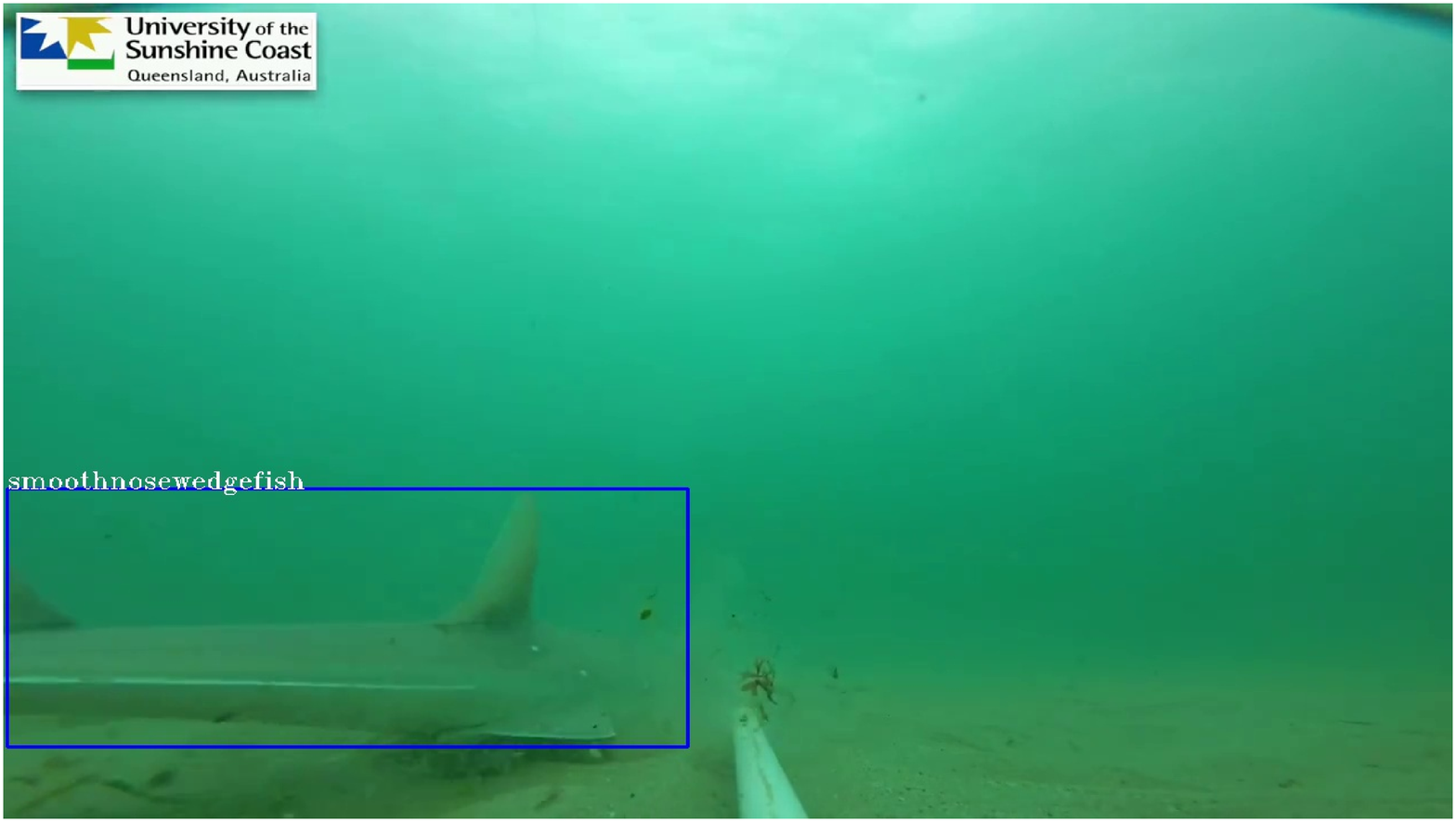}\\
        (a) & (b)\\
        \includegraphics[width=0.4\textwidth]{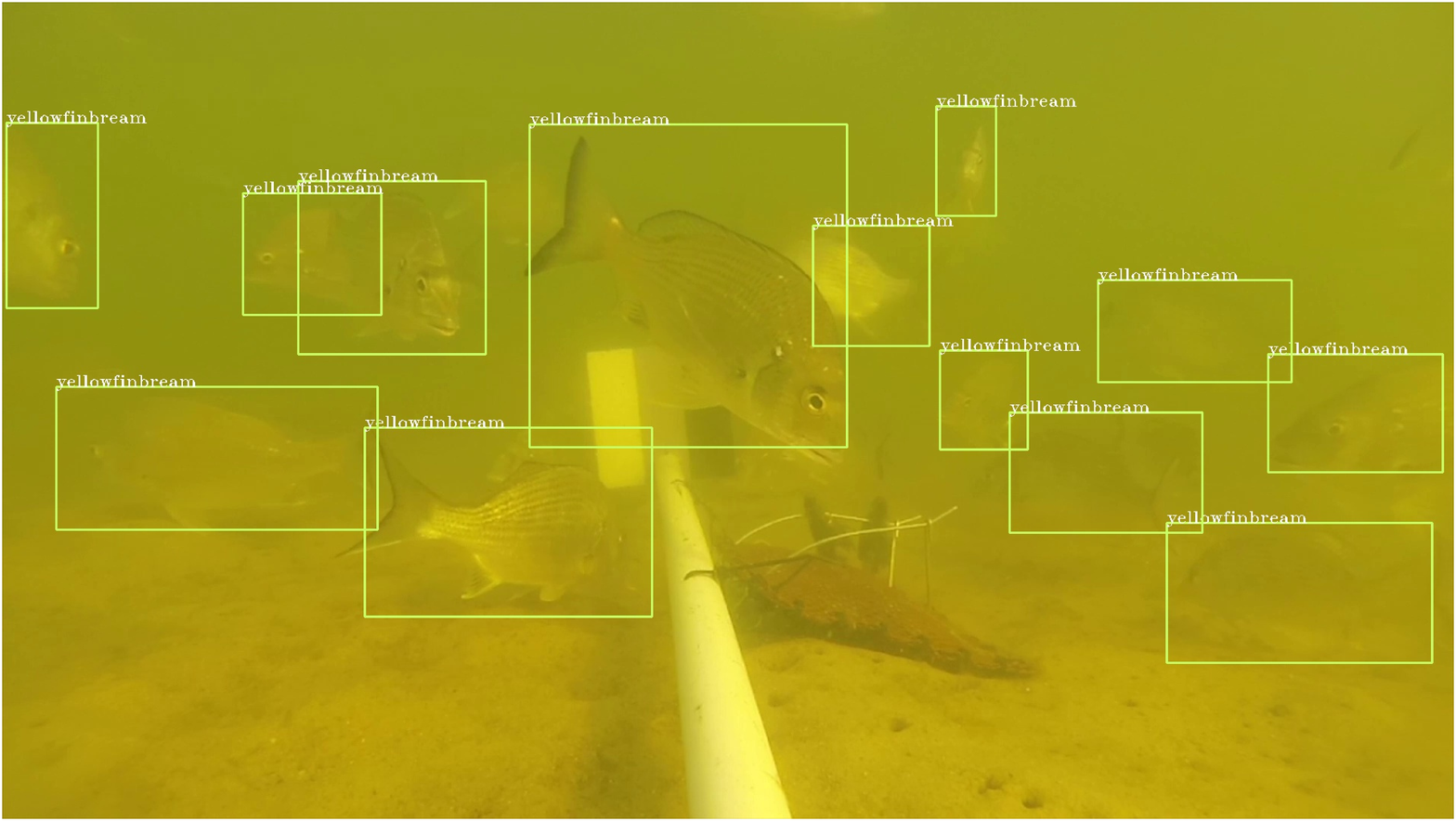}&
        \includegraphics[width=0.4\textwidth]{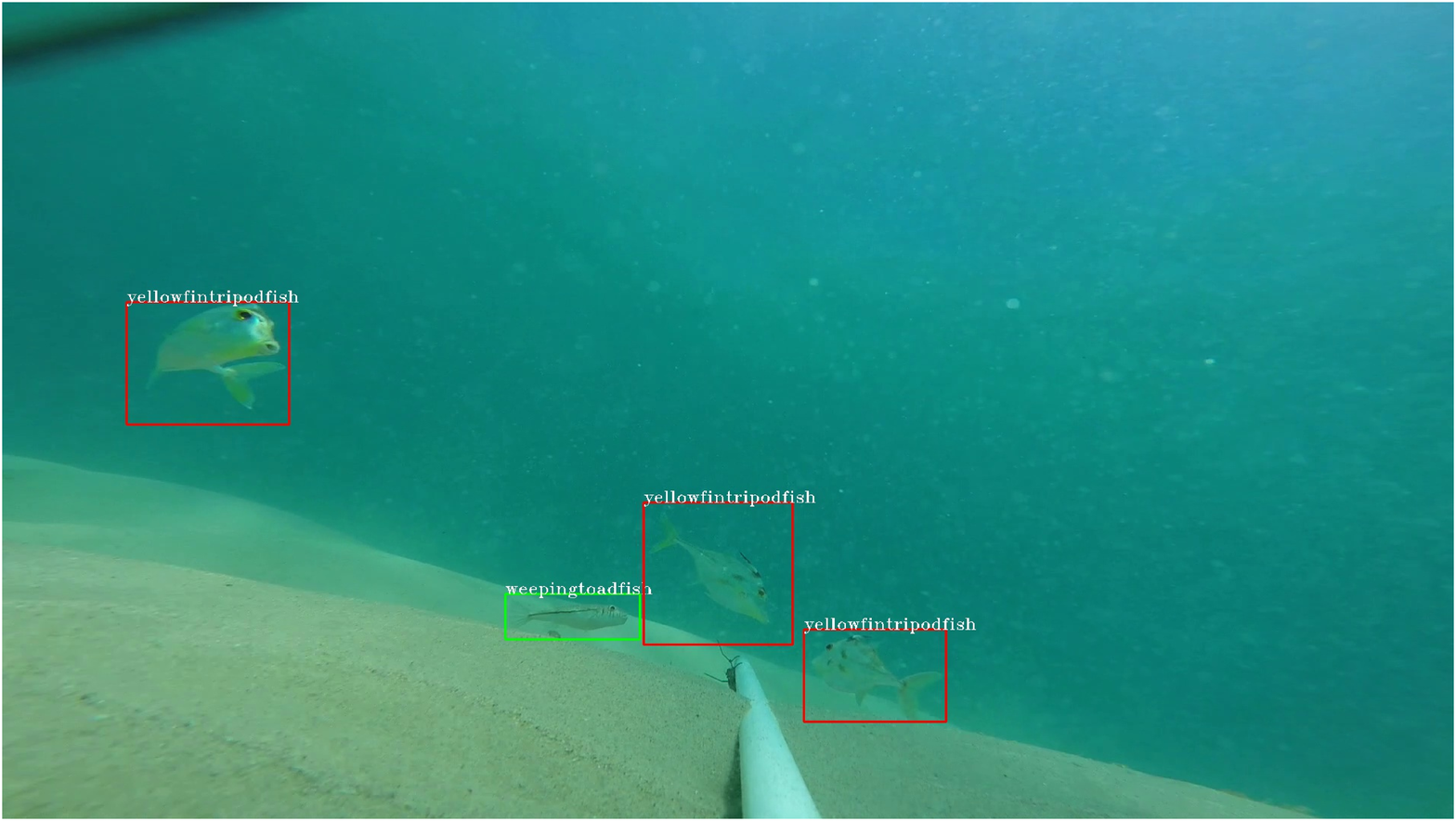}\\
           (c) & (d)\\
   \end{tabular}  \vspace{-2ex}
   \caption{Qualitative results on sample frames extracted from the underwater videos of  surf across southeast Queensland. Detected species are marked with a coloured bounding boxes.}
   \label{fig:FramedetectionResults}
   \vspace{-2ex}
\end{figure*}

\begin{figure*}[!htb]
   \centering
   \begin{tabular}{cc}
        \includegraphics[width=0.4\textwidth]{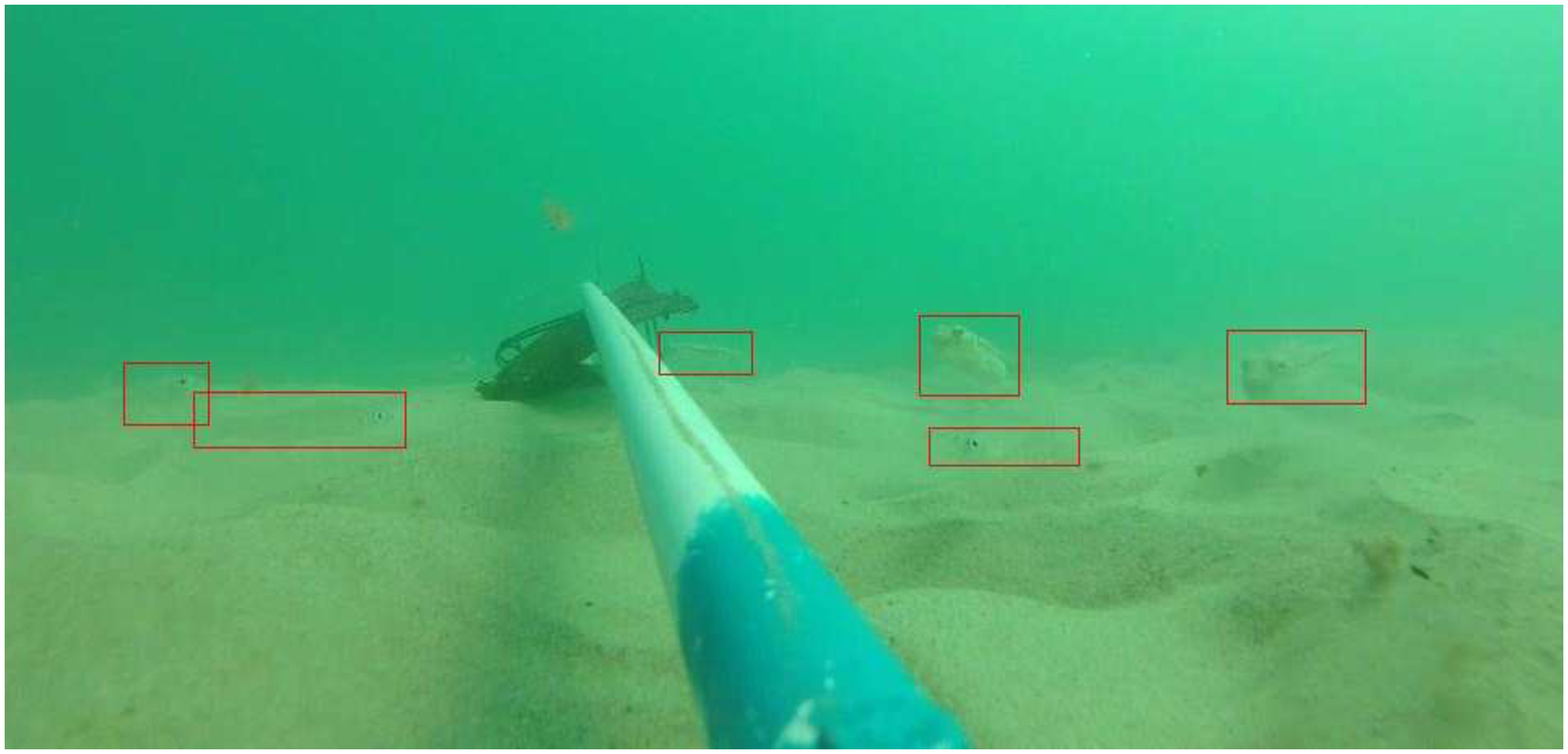}&
        \includegraphics[width=0.4\textwidth]{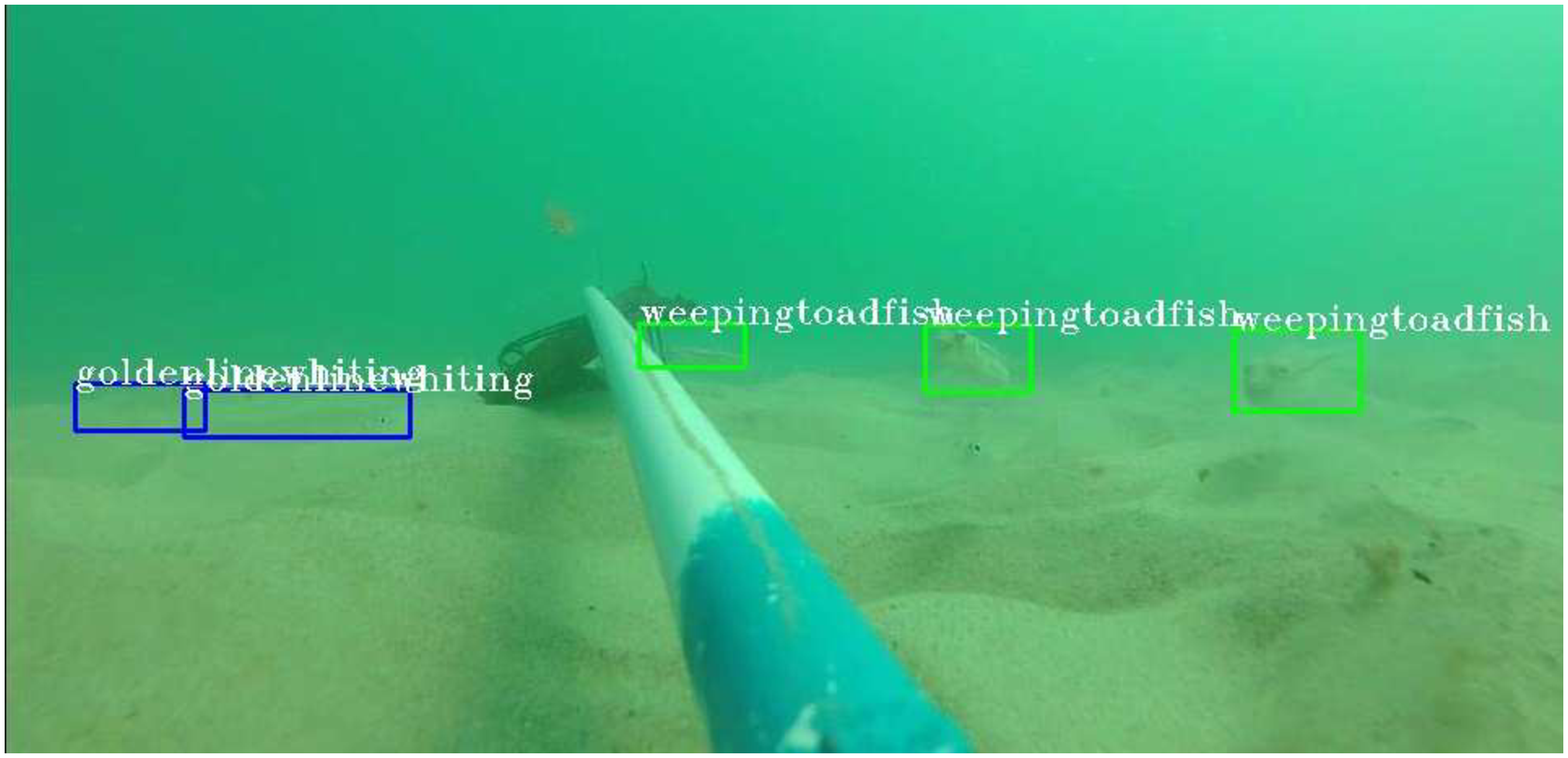}\\
        (a) & (b)\\
        \includegraphics[width=0.4\textwidth]{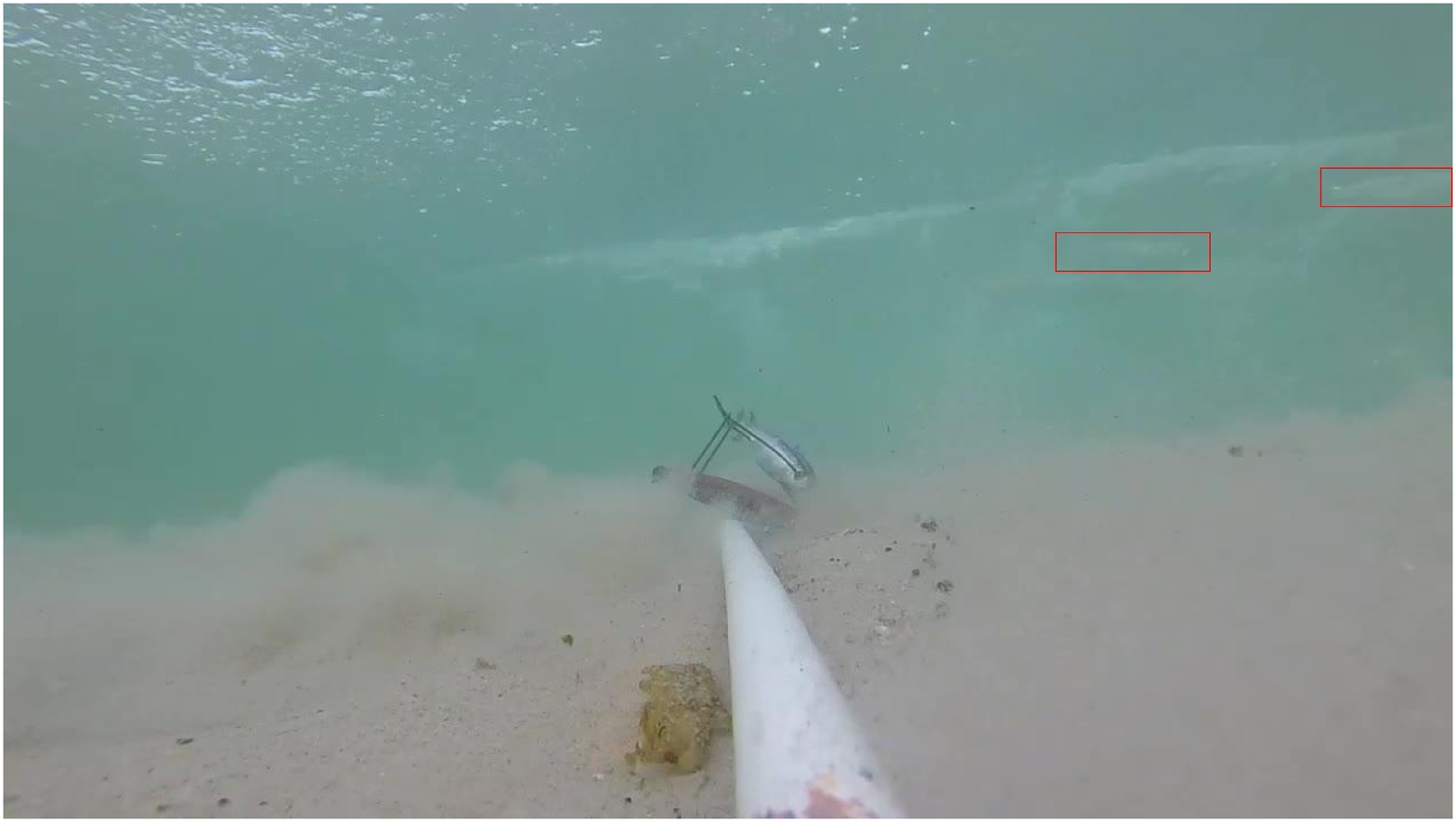}&
        \includegraphics[width=0.4\textwidth]{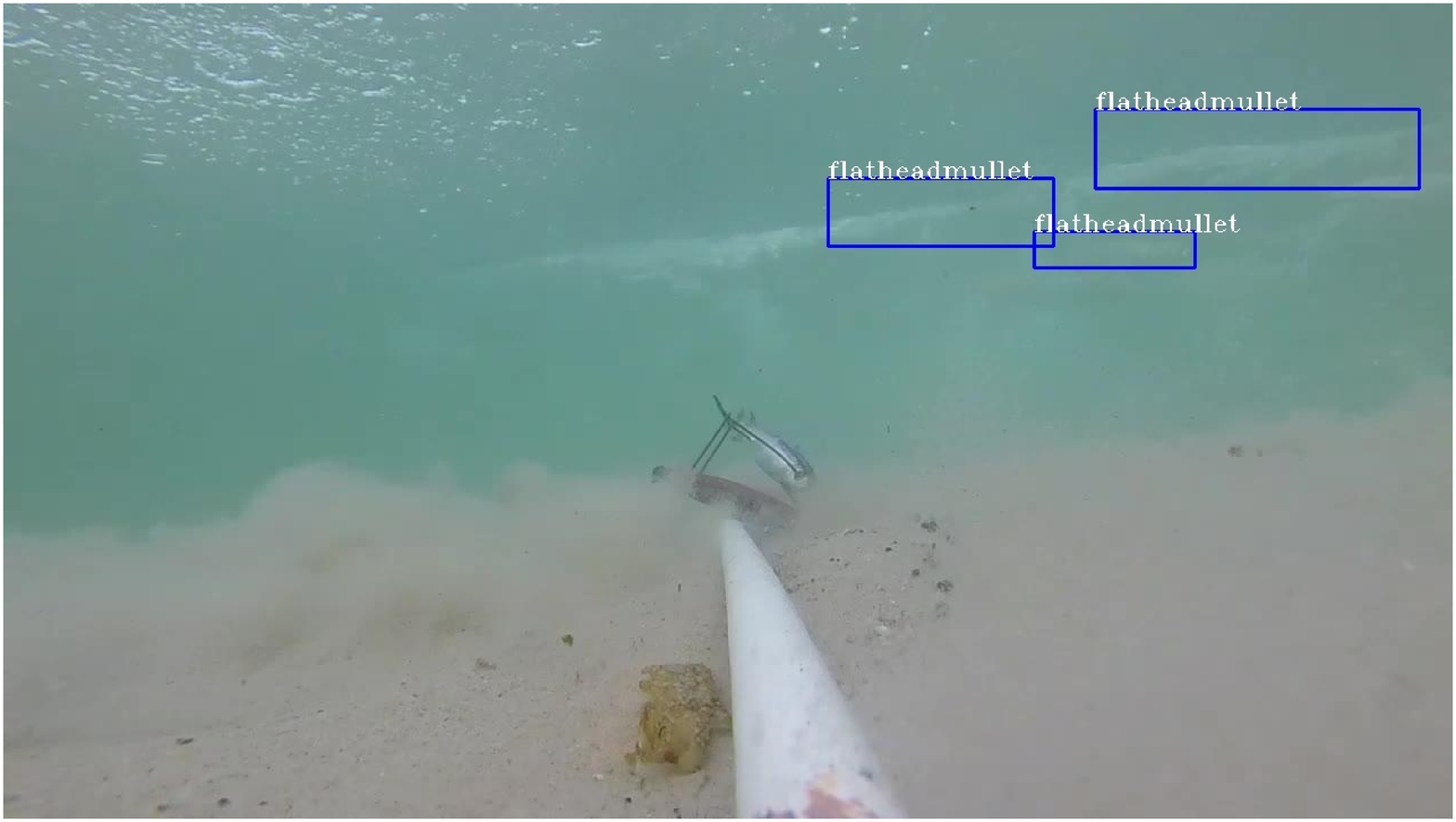}\\
        (c) & (d)\\
   \end{tabular}  \vspace{-1ex}
   \caption{Erroneous detections. (a,c) Ground Truth frames (b,d) Frames after detection.}
\label{fig:FramedetectionError}
\vspace{-3ex}
\end{figure*}

{\bf Detection results:} The detection results of several fish species from two sets of experiments are detailed in Table \ref{tab:surftableresults1} and  Table \ref{tab:surftableresults2} with the mean Average Precision (mAP) results. Table \ref{tab:surftableresults1} shows the results obtained from three different experiments using three network architectures considered in our experiments. The best result obtained among all iterations are presented here and the VGG-16 network outperformed. Mean AP of 0.72 and 0.71 and 0.71 were obtained after 70k iterations for VGG-16, CNN-M, and ZF respectively when the whole dataset was considered. However, accuracy was improved in experiment \rom{2} when species only have adequate training samples are considered. Table \ref{tab:surftableresults2} shows that maximum mAP of 82.4\% was achieved on the VGG16 network. An average time taken for processing an image for detection during testing process was 0.2 seconds (i.e. 5 fps) for VGG-16 and 0.1 seconds (i.e. 10 fps) for ZF and  CNN-M network models which imply that the system is capable of processing video in real-time.
%-----------------------Table1: Results surf dataset with 3 networks--------------------------------
\newcommand{\specialcell}[2][c]{  \begin{tabular}[#1]{@{}c@{}}#2\end{tabular}}
 \begin{table}[!htb]
 \vspace{-2ex}
\centering
\renewcommand{\arraystretch}{1.2}
\caption{Results obtained from the experiment \rom{1} on  fish  test dataset. AP represents average precision}
\vspace{-2ex}
\begin{tabular}{cccc}\hline
\bf {\specialcell{Species}} &\bf{\specialcell{AP on \\VGG-16}} &\bf{\specialcell{AP on \\CNN-M}} &\bf{\specialcell{AP on \\ZF}}\cr\hline
Mackerel tuna     &    1.00    &    1.00    &    0.55    \cr
Reticulated surf crab    &    1.00    &    1.00    &    1.00    \cr
School mackerel    &    1.00    &    1.00    &    0.55    \cr
Blueswimmer crab     &    0.91    &    0.90    &    0.91    \cr
Smooth flutemouth    &    0.91    &    0.81    &    0.91    \cr
Starry pufferfish     &    0.91    &    0.91    &    0.91    \cr
Sand crab     &    0.90    &    0.81    &    0.81    \cr
Spotted wobbegong     &    0.82    &  0.90    &    0.88    \cr
White spotted eagle ray    &    0.82    &    0.64    &    1.00    \cr
White spotted guitarfish    &    0.82    &    0.86    &    0.89    \cr    \hline
mAP   &    0.72    &    0.71    &    0.71    \cr    \hline
\end{tabular}
\label{tab:surftableresults1}
\vspace{-2ex}
\end{table}
%--------------------------------------------
% result obtained from seleted species
% -------------------------------------------
\begin{table}[!htb]
\centering
\renewcommand{\arraystretch}{1.2}
\caption{Results obtained from the experiment \rom{2} on fish test dataset. AP represents average precision}
\vspace{-2ex}
\begin{tabular}{cccc}\hline
%species    vgg16-90     vggm - 150    ZF-110
\bf {\specialcell{Species}} &\bf{\specialcell{AP on \\VGG-16}} &\bf{\specialcell{AP on \\CNN-M}} &\bf{\specialcell{AP on \\ZF}}\cr\hline
Bluespotted flathead     &1.000     &0.818    &0.831 \cr
Sand whiting  &1.000    &0.945    &0.909 \cr
Smooth golden toadfish     &1.000    &1.000     &1.000 \cr
Southern herring     &1.000    &0.947    &0.867 \cr
Smoothnose wedgefish     &0.996    &0.989    &0.892 \cr
Painted grinner    &0.986    &0.951    &0.972 \cr
Reticulate whipray    &0.909    &0.909    &0.909 \cr
Starry pufferfish     &0.909    &0.899    &0.807 \cr
Swallowtail dart    &0.909  &0.838    &0.994 \cr
Common stingaree    &0.906    &0.995    &0.97 \cr\hline
mAP    &0.824 &0.769    &0.750 \cr\hline
\end{tabular}
\label{tab:surftableresults2}
\vspace{-2ex}
\end{table}
%------------------------------------------------------------------------------------------
%% Line charts
\begin{figure}[!htb]
    \centering
      \includegraphics[width=0.4\textwidth]{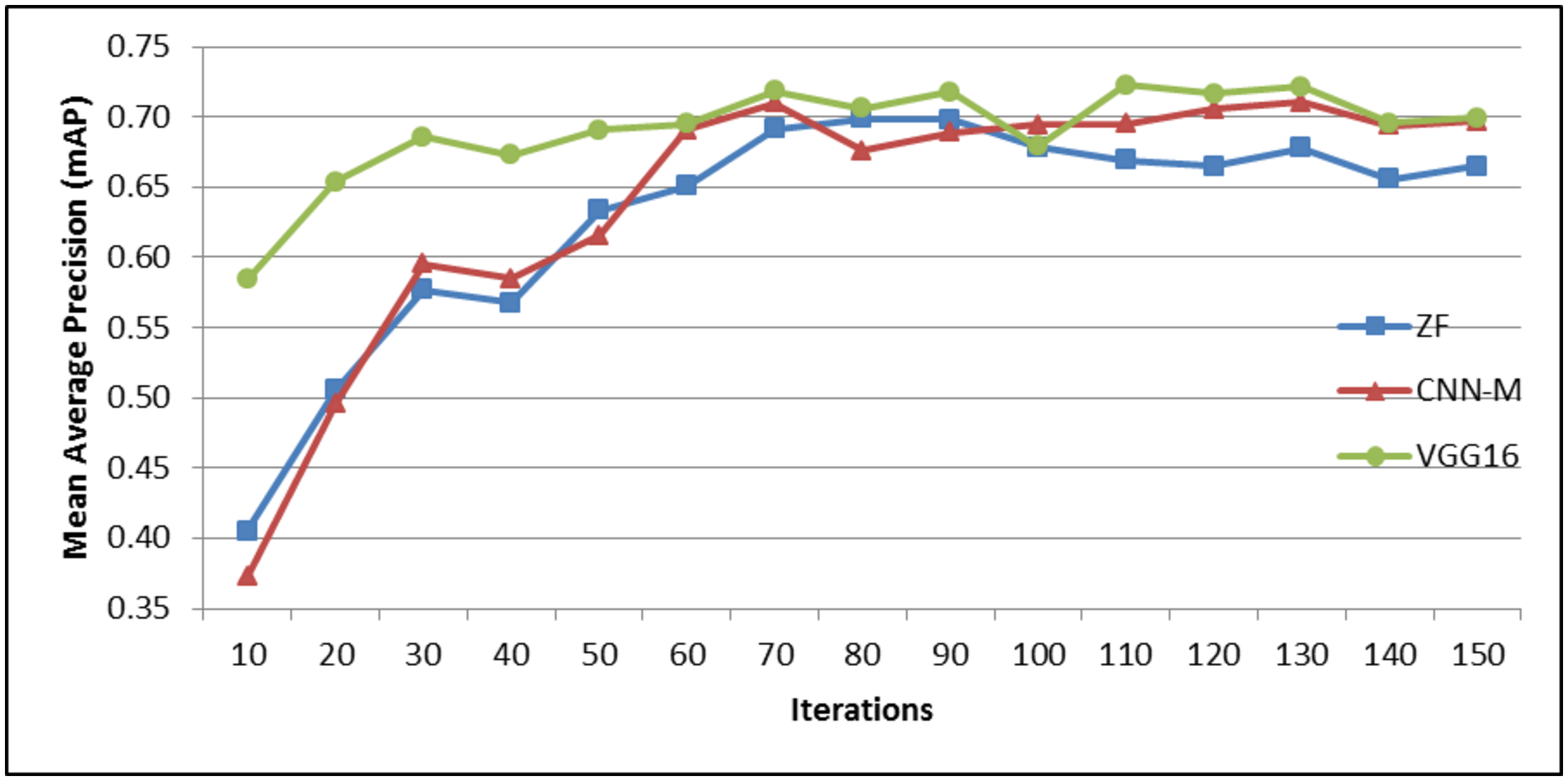}
      \vspace{-1ex}
    \caption{Mean average precision on test data using 3 different models from experiment set \rom{1}. X-axis represents iteration in thousands.}
   \label{fig:BarchartResults1}
   \vspace{-2ex}
\end{figure}
%------------------------------------------------------------
\begin{figure}[!htb]
   \centering
    \includegraphics[width=0.4\textwidth]{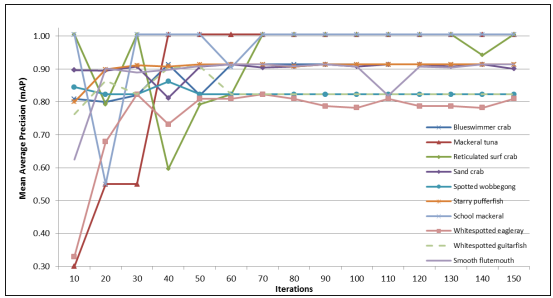}
    %\vspace{-1ex}
   \caption{Avg. precision analysis at different iterations on multiple fish and crustacean species in the dataset. X-axis represents iteration in thousands.}
   \label{fig:BarchartResults10species}
   \vspace{-3ex}
\end{figure}
%----------------------------------------------------------------------------------
\begin{figure}[!htb]
    \centering
      \includegraphics[width=0.4\textwidth]{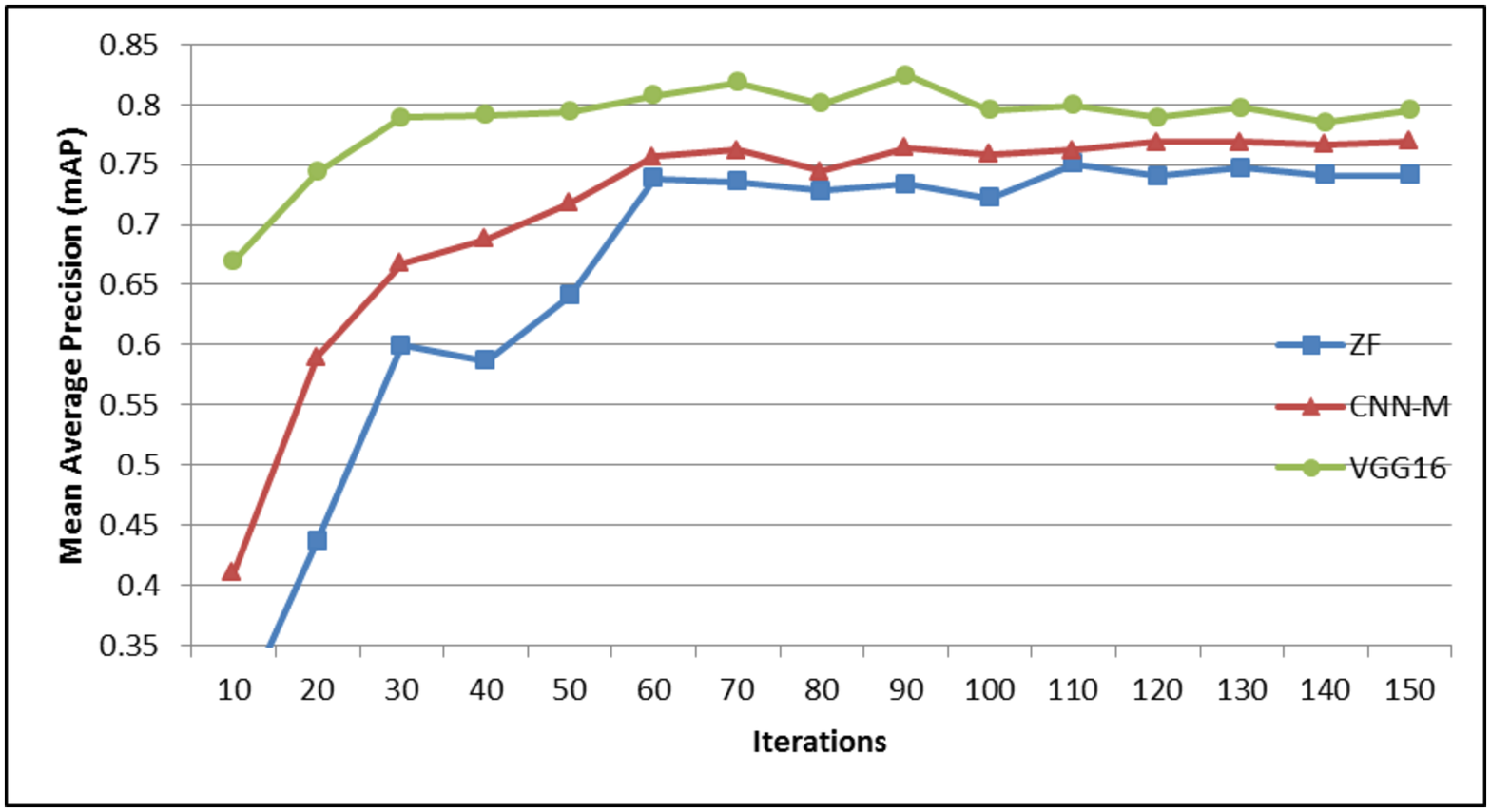}
      \vspace{-1ex}
    \caption{Mean average precision on test data using 3 different models from experiment set  \rom{2}. X-axis represents iteration in thousands.}
   \label{fig:BarchartResults2}
   \vspace{-3ex}
\end{figure}
%detail explanation of BAR charts
Fig. \ref{fig:BarchartResults1} shows how the mAP improves over iterations (x-axis represents iterations in thousands) during the testing process on three different network architectures and the highest mAP of 0.72 was obtained for VGG-16 at 70k iterations. The class-wise AP analysis has also presented for some sample species in Fig. \ref{fig:BarchartResults10species}.  Fig. \ref{fig:BarchartResults2} shows how the accuracies were improved over iterations in experiment \rom{2}. An mAP of 82.4\% was achieved on the test dataset after 90K iterations. 
The qualitative detection results of several sample frames are shown in Fig. \ref{fig:FramedetectionResults}. The detected region along with the species name is shown in all the detected frames. Some previous works on fish identification in the literature are significant as a fish classifier. However, our proposed system is more advanced as it detects the region of interest and classifies all the species in a single pipeline. As the existing works on fish identification were not conducted on any standard dataset and there is no public dataset available, a proper comparative study cannot be performed. However, Spampinato et al. \cite{Spampinato2016} reported an accuracy of 54\% on a dataset having only 10 species. 
An error analysis was performed on frames with incorrect detections. It was found that high levels of occlusion among a school of fishes was the main cause of the error.  Fig \ref{fig:FramedetectionError} shows some sample frames with incorrect detection. Two frames with ground-truthing and the same frames after detection are given side by side to aid understanding. Fig. \ref{fig:FramedetectionError}(b) shows one false negative case as the ground truth sample is occluded. Fig. \ref{fig:FramedetectionError}(d) shows two false positive cases as the pattern of ground truth fish data is identical with some background surf in this particular case.
%----------------------------------------------
%-------------------------------------------------------
\section{Conclusion}
\label{conclusion}
Automatic assessment of fish/species abundance using remote underwater video stream has tremendous potential over traditional approaches in terms of time and cost-effectiveness. The objective of the work was to develop a system for automatic detection and recognition of species from underwater videos. The significance of such a system has been studied and an appropriate work towards automation was not found in the literature on the assessment of fish abundance. An end to end deep learning approach was adapted to process a video stream and extract all the information required for the assessment. A range of experiments was conducted using different deep learning models and a comprehensive analysis of performance is presented. An mAP of 82.4\% was achieved across a very wide variety of marine species.
 %--------------------------------------------------------------------------
The main contributions of our work are, therefore:
\begin{itemize}
\item Proposed a high-performance fish identification system by fine-tuning the `Faster R-CNN' which has been adapted to our problem 
\item Presentation of a wide range of experiments for underwater fish detection and identification using three different (small, medium and large sizes) state-of-the-art classification network models
\item Introduction of a newly developed fish abundance dataset which contains 50 different species from multiple beaches and estuarine sites across southeast Queensland, Australia. The number of species considered in these experiments is significantly higher than previously proposed approaches. The dataset is annotated and standardised in PASCAL VOC \cite{Everingham15} format using the Vatic video annotation tool \cite{springerlink:10.1007/s11263-012-0564-1}.
\end{itemize}
In future, we aim to further improve the performance by enhancing the CNN architecture and training the system with more samples in the training dataset.
\section*{Acknowledgment}
This research was partly funded by the National Environment Science Program (NESP) Tropical Water Quality Hub Project No 3.2.3. Videos are made available through a collaboration between researchers at University of Sunshine Coast and Griffith University.
\bibliographystyle{IEEEtran}      % mathematics and physical sciences
\bibliography{Fishdetection}   % name your BibTeX data base
\end{document}